\pdfoutput=1

\documentclass[11pt]{article}

\usepackage{acl}

\usepackage{times}
\usepackage{latexsym}
  \usepackage[frozencache=true,cachedir=minted-cache]{minted} 
\usepackage{adjustbox}

\usepackage[T1]{fontenc}

\usepackage[utf8]{inputenc}

\usepackage{microtype}

\usepackage{inconsolata}

\usepackage{amssymb}
\usepackage{amsmath}
\usepackage{enumitem}
\usepackage{graphicx}
\graphicspath{ {./images/} }

\usepackage{tabularx}
\usepackage{multirow}
\usepackage{dsfont}
\usepackage{url}
\usepackage{xfrac}
\usepackage{float}

\usepackage{pdfpages}

\usepackage{todonotes} 

\title{\textsc{MDace}: MIMIC Documents Annotated with Code Evidence}

\author{
Hua Cheng\textsuperscript{1}, Rana Jafari\textsuperscript{1}, April Russell\textsuperscript{1}, Russell Klopfer\textsuperscript{1}, \\ 
{\bf Edmond Lu\textsuperscript{1}, Benjamin Striner\textsuperscript{1}, Matthew R. Gormley\textsuperscript{2}}  \\
\textsuperscript{1}3M Health Information Systems, \textsuperscript{2}Carnegie Mellon University\\
\small{\texttt{\{hcheng, rjafari, arussell3, rklopfer, elu3, bstriner\}@mmm.com, mgormley@cs.cmu.edu}}
}

\begin{document}
\maketitle

\begin{abstract}
    We introduce a dataset for evidence/rationale extraction on an extreme multi-label classification task over long medical documents.
    One such task is Computer-Assisted Coding (CAC) which has improved significantly in recent years thanks to advances in machine learning technologies. However, simply predicting a set of final codes for a patient encounter is insufficient, as CAC systems are required to provide supporting textual evidence to justify the billing codes. A model able to produce accurate and reliable supporting evidence for each code would be a tremendous benefit. However, a human-annotated code evidence corpus is extremely difficult to create because it requires specialized knowledge. In this paper, we introduce \textsc{MDace}, the first publicly available code evidence dataset, which is built on a subset of the MIMIC-III (English) clinical records. The dataset -- annotated by professional medical coders -- consists of 302 Inpatient charts with 3,934 evidence spans and 52 Profee charts with 5,563 evidence spans. 
    We implemented several evidence extraction methods based on the EffectiveCAN model \citep{liu-etal-2021-effective} to establish baseline performance on this dataset. \textsc{MDace} can be used to evaluate code evidence extraction methods for CAC systems, as well as the accuracy and interpretability of deep learning models for multi-label classification. We believe that the release of \textsc{MDace} will greatly improve the understanding and application of deep learning technologies for medical coding and document classification.
\end{abstract}

\section{Introduction}
\label{sec:intro}


In extreme multi-label text classification (XMLTC) a document is assigned a small number of labels from an extremely large set of possible labels. This large label space poses a challenge for machine learning (ML) which is compounded by the length of input seen in long-document classification. While there is a wide range of document classification datasets, only a limited number of those contain rationales or evidence associated with the labels. Of those that do, none (as of writing) are in the extreme multi-label classification setting or apply to long-documents. We present a new dataset for evidence extraction on long documents in an extreme multi-label classification setting. We also provide benchmark results using established techniques using neural networks.

Computer-Assisted Coding (CAC) is a real world XMLTC application that uses natural language processing (NLP) techniques to extract procedure and diagnosis codes from the documentation of patient encounters. MIMIC-III (Medical Information Mart for Intensive Care) \citep{mimic} is an open-access dataset comprised of hospital records associated with patients admitted to the critical care units of the Beth Israel Deaconess Medical Center. For each patient record/chart, the data related to billing includes diagnostic codes, procedure codes, clinical notes by care providers (discharge summaries, radiology and cardiology reports, nursing notes, etc., all in English), and other patient demographic data. The MIMIC records were originally coded with the alphanumeric code system ICD-9 (International Classification of Diseases) \citep{world1978international}, which contains approximately 14,000 codes overall.

Since the release of MIMIC-III, there has been a surge of research on using ML models to predict billing codes based on the clinical text \citep{ji-etal-2022-unified}. However, the MIMIC database does not contain the association between the billing codes and the clinical notes, i.e., the specific narratives in the notes supporting the codes are not present. CAC systems are required to extract text evidence (i.e. rationales) to support the generated billing codes. There is no dataset for reference code evidence as it requires medical coding expertise and is costly to build. As a result, work until this point can only illustrate qualitatively that their models can extract text evidence that looks reasonable to humans. This approach is time-consuming and makes the comparison of different methods extremely difficult. The need for a reference evidence dataset is obvious.

In many parts of the world, the ICD-9 code system is out of date. Most countries are currently using the much more robust and specific alphanumeric code system, ICD-10 \citep{world2004international}. The U.S. version, ICD-10-CM, has approximately 69,000 codes, while the procedures (PCS) have about 82,000 codes. 
Only documents generated as a result of a face-to-face visit with an allowable provider should be reviewed for direct ICD-10 code abstraction. This includes Progress Notes, History and Physicals, Consults and Operatives Notes, etc., and excludes nursing notes. For procedure code selection, only a procedure or operative note is acceptable. 
For these reasons, the ML models trained on the MIMIC-III discharge summaries to predict ICD-9 codes have limited value for medical coding in reality. MIMIC-IV \citep{Johnson2020, johnson2023mimic} improved upon MIMIC-III in many ways, one of which is the addition of ICD-10 codes. But at the time of our annotation project, the clinical notes associated with the patient records had not been released. 

In this paper, we introduce \textsc{MDace}, the first publicly available code evidence dataset\footnote{The dataset and software (under the MIT license) are available at https://github.com/3mcloud/MDACE/.} built on a subset of the MIMIC-III clinical records. The dataset contains evidence spans for diagnosis and procedure codes annotated by professional medical coders. Each span contains the billing code and the text offsets in the respective clinical note. We provide Python scripts for merging our evidence representation with the MIMIC NOTEEVENTS table to obtain the true evidence so as to comply with \emph{The PhysioNet Credentialed Health Data License}. To broaden its use, we automatically map between ICD-10 and ICD-9 codes so that the evidence can potentially be used with the MIMIC-IV corpus. \textsc{MDace} addresses a critical need for the automatic evaluation of code evidence generated by CAC models as well as the rationales extracted by XMLTC systems.


\section{Related Work} 
\label{sec:literature}

\begin{table*}[t]
\small
\centering
\begin{tabular}{lccccc}
\hline
\textbf{Dataset} & \textbf{Avg. Tokens} & \textbf{Tot. Labels} & \textbf{Tot. Classes} & \textbf{Avg. Labels} & \textbf{Avg. Evidence} \\
\hline
\textsc{MDace} (IP) & 19,372 & 918 & 2 & 11.30 & 13.03 \\
\textsc{MDace} (Profee) & 11,116 & 652 & 2 & 31.35 & 106.98 \\
EvidenceInference \citep{lehman-etal-2019-inferring} & ~4,200 & 1 & 3 & 4.19 & 4.19 \\
MovieReview \citep{zaidan-etal-2007-using} & 774 & 1 & 2 & 1 & 11.36 \\
FEVER \citep{thorne-etal-2018-fever} & 327 & 1 & 3 & 1 & 1.77 \\ 
e-SNLI \citep{NEURIPS2018_4c7a167b} & 16 & 1 & 3 & 1 & 1 \\
\hline
\end{tabular}
\caption{Comparison of a sampling of classification datasets that have evidence annotations (i.e. rationales) in terms of the average number of tokens per document, total number of unique labels / classes, average number of labels per document (i.e. for standard classification tasks this is 1, for multi-label settings this is $>1$), and average number of evidence annotations (i.e. highlights) per document. Our new \textsc{MDace} dataset consists of two parts: Inpatient (IP) and Profee. }
\label{tab:datasets}
\end{table*}

With the recent increased attention to the interpretability of deep learning models, datasets containing explanations in different forms (highlights, free-text, structured) have been curated. \citet{wiegreffe-2021} provide a list of 65 datasets for various explainable NLP tasks, and \citet{feldhus2021ther} present the results of different explanation generation models trained on these datasets. 

The primary differences between \textsc{MDace} and existing explanation/evidence/annotator rationale datasets for classification tasks are illustrated in Table~\ref{tab:datasets}. Prior datasets focused on shorter documents (except for EvidenceInference \citep{lehman-etal-2019-inferring}), and the tasks usually involve annotators highlighting evidence that supports or refutes a single claim \citep{lehman-etal-2019-inferring, zaidan-etal-2007-using, thorne-etal-2018-fever}. In contrast, our task is an extreme multi-label classification problem: a medical coder must find multiple codes (i.e. labels) from a large target set of codes based on the documentation while highlighting one or more pieces of evidence for each label. 
To the best of our knowledge, \textsc{MDace} is the only publicly available dataset with evidence annotations for long documents in an extreme multi-label classification setting. 

Many private datasets have been developed for evidence extraction for medical coding, e.g., \citet{sen2021}.  \citet{DeYoung2022} described a MIMIC-III subset annotated with potential evidence spans and assigned a ranked list of ICD-10 codes. However, these datasets are not publicly available and cannot be used to improve research on evidence extraction. In addition, \textsc{MDace} was created with an annotation process closely mimicking coding in a professional setting. Coders reviewed and annotated charts containing multiple clinical notes instead of individual unrelated notes, and coded both procedure and diagnosis codes. There also exists automatically created datasets, for example, \citet{searle-etal-2020-experimental} used a semi-supervised approach to create a silver-standard dataset of clinical codes from only the discharge diagnosis sections of the MIMIC-III discharge summary notes, with a small sample validated by humans.  


There has been a surge in neural network models for automatic medical coding in the past several years. \citet{mullenbach2018explainable} first introduced a convolutional neural net with an attention mechanism, where the label dependent attention weights were used as token importance measure for the model interpretability. 
\citet{liu-etal-2021-effective} extended this work by incorporating the squeeze-and-excitation network \citep{hu-shen-2018-squeeze} into the text encoder to obtain better contextual text representations.
\citet{xie-etal-2019-cikm} used the multi-scale convolutional attention while \citet{vu-etal-2020-ijcai} proposed to combine Bi-LSTM and an extension of structured self-attention mechanism for ICD code prediction. Some other recent models that achieved state-of-the-art results on the MIMIC-III full code set include \citet{kim-etal-2021-read, Hu-Teng-2021, yuan-etal-2022-code}. 
There are also a large number of Transformer based models for medical coding, e.g., \citet{Liu2022hier, pascual2021}, but they often only predict the top 50 codes. One exception is PLM-ICD \citep{huang2022plm}, which used domain-specific pretraining, segment pooling, and label-aware attention to tackle the challenges of coding and improve performance. 

Many of the above works are able to use the attention weights to identify the text snippets that justify code predictions. But there is no quantitative evaluation of the quality of the snippets mostly due to the lack of reference evidence. 

Works that use semi-supervised learning for explanation tasks in NLP more broadly include \citet{zhong2019fine, pruthi-etal-2020-weakly, segal-etal-2020-simple}, where \citet{segal-etal-2020-simple} used a linear tagging model for identifying answer snippets in question answering. Although they are not directly related to medical coding, we can apply their approaches for evidence extraction with the help of the \textsc{MDace} dataset.





\section{Challenges and Solutions}
\label{sec:challenge}

MIMIC-III poses a number of challenges for creating a reference code evidence dataset. These challenges include the different coding specialties (Inpatient \& Profee) and code systems (ICD-9, ICD-10 \& CPT). This section discusses these challenges and describes our process to increase the usability of \textsc{MDace}.

\subsection{Coding Specialties}
\label{sec:ip-pf}

MIMIC-III contains both ICD-9 codes, which are used for inpatient coding, and CPT (Current Procedure Terminology) codes, which are maintained by the American Medical Association (AMA) and used for outpatient facility and professional fee (Profee) billing in the U.S. (See Appendix~\ref{app:code-term} for details). There are approximately ten thousand CPT-4 codes. It was necessary to have different coders for each of these tasks (Inpatient vs. Profee) because it is unusual that one person be experienced in both areas. This means that inpatient coders tend to be more skilled ICD coders, while Profee coders are often skilled CPT coders within their domain. ICD codes are also applied to Profee charts to meet medical necessity requirements which ensure that the patient’s bill is paid by insurance companies.   

For this reason, we hired two coding teams with two professional coders each for Inpatient and Profee coding. Although both teams coded diagnosis codes, the actual codes can be different due to different coding rules.

For either coding scenario, a coder usually looks for sufficient evidence that supports a code and ignores equally good evidence that she comes across later to save time. This poses a challenge for evaluating CAC systems which can generate multiple pieces of evidence for a code that may or may not overlap with the \emph{sufficient} reference evidence. To overcome this challenge but still finish the annotations in a reasonable time frame, we asked our coders to annotate sufficient evidence for Inpatient coding but \emph{complete} evidence for Profee coding.

\subsection{Code Mappings}

Since ICD-9 coding has been discontinued, updating the MIMIC-III dataset with ICD-10 codes and evidence will benefit research that targets real-world coding problems. \textsc{MDace} is designed to contain evidence for both ICD-9 and ICD-10 codes so that it can be used to evaluate evidence extraction of CAC models trained on either MIMIC-III or MIMIC-IV.

We chose to use ICD-10 for annotation because, firstly, most coders are more familiar with the ICD-10 code system, and secondly, ICD-10 codes are more specific, so the mapping from an ICD-10 code to ICD-9 is less ambiguous than the other way around. Our coders annotated a subset of the MIMIC-III charts with ICD-10 codes and their evidence, which were then automatically mapped to ICD-9 through the General Equivalence Mappings (GEMs)\footnote{GEMs are a comprehensive translation dictionary developed by multiple health organizations in the U.S. to effectively translate between the ICD-9 and ICD-10 codes.} \citep{CMS:2009}. GEMs contain six types of mappings, including Identical match, Approximate match, Combination map, and No Map, etc. To ensure the quality of code mapping, we follow the procedure in Appendix~\ref{app:codeMap} to backward map ICD-10 to ICD-9. This process allows all annotated ICD-10 codes to be mapped except for two in our dataset.








\subsection{Annotation Workflow}

Medical coding is an extremely complex task, and there is often disagreement among coders. Given the large number of notes and codes in each MIMIC-III record \citep{Su2019read}, it is impractical for our coders to first decide the best ICD-10 code for a MIMIC ICD-9 code and then annotate the narrative evidence in clinical notes for that code. Therefore, our coders followed their natural workflow of coding each chart from scratch. However, the original MIMIC codes and their possible ICD-10 mappings were made available to them. If there were MIMIC codes unaccounted for after completing a chart, those could be used as reference to re-review the chart and annotate accordingly. If the coders could not find evidence after reviewing again -- for example, if the required note was missing -- they simply made a note in their coding reports.

We used a tool called INCEpTION \citep{tubiblio106270} to help our coders to review and annotate MIMIC charts. This tool allows them to browse through the clinical notes, highlight text spans and assign labels (billing codes) to the spans. The annotation guideline is illustrated in Appendix~\ref{app:guide}.

We selected charts from \citet{mullenbach2018explainable}'s \emph{test} set to be annotated by our specialists. Batches of 50 charts were chosen at random. For each batch, all eligible documents were extracted, not just discharge summaries. Our coders worked on one batch at a time. The project lasted two months.

\subsection{Inter-Annotator Agreement}

As the first step of the annotation process, we measured the inter-annotator agreement to assess the reliability of the annotations. To quantify the quality of annotations, two coders independently annotated sufficient (for Inpatient) or complete (for Profee) evidence for the same three charts, and we measured the inter-coder agreement. Next, they reviewed each other's annotations where they disagreed to investigate the reasons for disagreement and determine if they could reach an agreement. If they still disagreed, their supervisor made the final call. Once all disagreements were resolved, the coders started working on the first batch of charts following the same coding practice.

We used Krippendorf’s $\alpha$ \citep{krippendorff2004content} as an agreement measure, as it allows for assigning multiple labels to a span, which is the case in medical coding. The agreement for initial and final coding is given in Table \ref{table:inter-agreement}, where the $\alpha$ values higher than $0.80$ could be interpreted as strong agreement. Two other agreement measures, Fleiss $\kappa$ \citep{fleiss1981stat}, and Hooper's measure of indexing consistency \citep{funk1983indexing}, are also reported. Punctuation was disregarded in these calculations. 
 
We observed two sources that accounted for the low initial agreement. One source is that the coders annotated the same or similar evidence from different locations in the same chart. The other source of disagreement came from external cause codes and symptom codes, which are not essential for billing, so some coders chose to code them while others did not. For Profee coding, the initial disagreement was also due to the lack of experience of one coder. Examples of these disagreements are given in Appendix~\ref{app:disagree}. These cases were resolved in the re-review process, and should be treated as agreements. After the review process, the inter-annotator agreement is high for both Inpatient and Profee coding.


\begin{table}[t]
\small
\centering
\begin{tabular}{lccc}
\hline
 & \textbf{Inpatient} & \textbf{Profee}\\
\textbf{Number of Annotations} & 384 & 1,282 \\
\noalign{\global\arrayrulewidth=1.pt}
\hline
\textbf{Agreement on Initial Annotations} & & \\
\hline
Krippendorf’s $\alpha$  & {$0.53$} & {$0.24$}  \\
\noalign{\global\arrayrulewidth=0.25pt}
\hline
Fleiss' $\kappa$ & {$0.53$} &  {$0.24$} \\
\hline
Hooper’s Measure & {$0.65$} & {$0.38$}  \\
\noalign{\global\arrayrulewidth=1.pt}
\hline
\textbf{Agreement after Review} & & \\
\hline
Krippendorf’s $\alpha$ & {$0.97$} & {$0.96$}  \\
\noalign{\global\arrayrulewidth=0.25pt}
\hline
Fleiss' $\kappa$ & {$0.97$} & {$0.96$}  \\
\hline
\end{tabular}
\caption{Inter-annotator agreement measures on initial and reviewed annotations}
\label{table:inter-agreement}
\end{table}


\section{Dataset Analysis}
\label{sec:stats}

In this section, we present various statistics of \textsc{MDace}, including the number of annotated charts, documents, unique codes, and evidence spans (Table \ref{tab:overall-count}). Since annotating complete evidence is more time-consuming than annotating sufficient evidence, the Profee coders only completed a small subset (52) of the 302 Inpatient charts.

\begin{table}[t]
\small
\centering
\begin{tabular}{lll}
\hline
\textbf{Annotated} & \textbf{Inpatient} & \textbf{Profee}\\
\hline
Encounters & 302 & 52 \\
Documents & 604 & 588 \\
ICD-9 Codes & 918 & 652  \\ 
ICD-10 Codes & 1,024 & 734  \\ 
Evidence for ICD-9 & 3,934 & 5,563 \\
Evidence for ICD-10 & 3,936 & 5,563 \\
Average evidence length (tokens) & 2.18 & 1.96 \\
\hline
\end{tabular}
\caption{Summary of \textsc{MDace} (Profee code and evidence counts include CPT codes)}
\label{tab:overall-count}
\end{table}


Tables \ref{tab:ip-doc-evidence} shows the distribution of evidence spans in different note categories. Research on deep learning models for CAC has been mostly focused on using discharge summaries for code prediction. The tables show that although discharge summaries capture the majority of coding related narratives for Inpatient, they are insufficient for Profee coding. Other notes, such as Physician and Radiology notes, should also be used.

Table \ref{tab:code-overlap} shows the overlap between the MIMIC codes and \textsc{MDace} codes\footnote{We ignored CPT codes for Evaluation and Management (E\&M), which are in the range of 99201 and 99499 as they require a decision making calculator to arrive at the correct CPT codes rather than simply depending on the clinical text.}. There is less than $50\%$ code overlap, indicating that a high percentage of MIMIC codes are missing from our annotations. There are two possible explanations for this: firstly, over 37\% of the 302 MIMIC encounters are missing operative notes, and as a result, the coders could not annotate the procedure codes accounting for 33\% of the missing Inpatient codes; and secondly, coding guidelines have changed over the years, and our coders were likely following different coding standards from the MIMIC coders.
However, verifying such a claim without information about the MIMIC coding process is impossible. It should be noted that a similar observation of low agreement with MIMIC coders based on 508 re-annotated discharge summaries was also reported in \citep{kim-etal-2021-read}.
Our coders added an average of 25 extra codes per chart for Profee coding because of their effort to annotate all evidence spans. The final codes of the annotated charts consist of the original MIMIC codes and extra codes added through annotation. Only codes verified by our annotators have related evidence.


\begin{table}[t]
\small
\centering
\begin{tabular}{llll}
\hline
\textbf{} & \textbf{Note Category} & \textbf{Evidence} & \textbf{Percentage}\\
\hline
   & Discharge Summary & 3,434 & 87.3\\
   & Physician & 364 & 9.3  \\ 
IP & Radiology & 60 & 1.5 \\
   & General & 28 & 0.7 \\
   & Nutrition & 19 & 0.5 \\
\hline
   & Physician & 2,082 & 37.4 \\
   & Discharge Summary & 1,584 & 28.5 \\
PF & Radiology & 1,269 & 22.8 \\
   & ECG & 256 & 4.6 \\
   & Echo & 207 & 3.7 \\
   & Rehab Services & 66 & 1.2 \\
\hline
\end{tabular}
\caption{Distribution of evidence spans in Inpatient and Profee notes (cutoff at 10)}
\label{tab:ip-doc-evidence}
\end{table}

\begin{table}[t]
\small
\centering
\begin{tabular}{lcc}
\hline
\textbf{Codes} & \textbf{Inpatient} & \textbf{Profee}\\
\hline
MIMIC & 5,250 & 694 \\
\textsc{MDace} & 3,414 & 1,630 \\
Agreed & 2,370 (45.1\%) & 306 (44.1\%) \\ 
Missed & 2,880 (54.9\%) & 388 (55.9\%) \\
Added (average) & 3.457 & 25.462 \\
\hline
\end{tabular}
\caption{Comparison of MIMIC-III and \textsc{MDace} codes}
\label{tab:code-overlap}
\end{table}

Table~\ref{tab:map-type} summarizes the mapping from ICD-10 to ICD-9 codes. The majority of the mappings, 92\% for Inpatient and 87\% for Profee, were either verified by coders during the annotation process or based on a single identical or approximate match in GEMs. This gives us high confidence in the quality of the mapped ICD-9 codes.

\begin{table}[t]
\small
\centering
\begin{tabular}{lll}
\hline
\textbf{ICD-10 to ICD-9} & \textbf{Inpatient} & \textbf{Profee}\\
\hline
Coder Verified & 2,525 (64.2\%) & 1,606 (28.9\%) \\
Identical match & 417 (10.6\%) & 1,387 (24.9\%) \\
Approximate match & 687 (17.5\%) & 1,847 (33.2\%) \\ 
Multiple match & 244 (6.2\%) & 704 (12.6\%) \\ 
Other & 61 (1.6\%) & 19 (0.3\%) \\
\hline
\end{tabular}
\caption{Distribution of code mappings}
\label{tab:map-type}
\end{table}

\section{Evidence Extraction Methods}
\label{sec:methods}

This section introduces several evidence extraction methods that we implemented within a convolutional neural network based model to establish baselines for code evidence extraction on \textsc{MDace}.

\subsection{EffectiveCAN}

EffectiveCAN \citep{liu-etal-2021-effective} is a convolution-based multi-label text classifier that achieved state-of-the-art performance on ICD-9 code prediction on MIMIC-III. It encodes the input text through multiple layers of residual squeeze-and-excitation (Res-SE) convolutional block to generate informative representations of the document. It uses label-wise attention to generate label specific representations, which has been widely used to improve predictions as well as to provide an explanation mechanism of the model, e.g., \citep{mullenbach2018explainable}. We chose EffectiveCAN as our base model for its simplicity, efficiency, and high performance. Its attention weights can be viewed as soft masks, making it a natural fit for producing baseline evidence results on \textsc{MDace}.

\begin{figure*}[ht]
    \centering
    \includegraphics[width=1\textwidth]{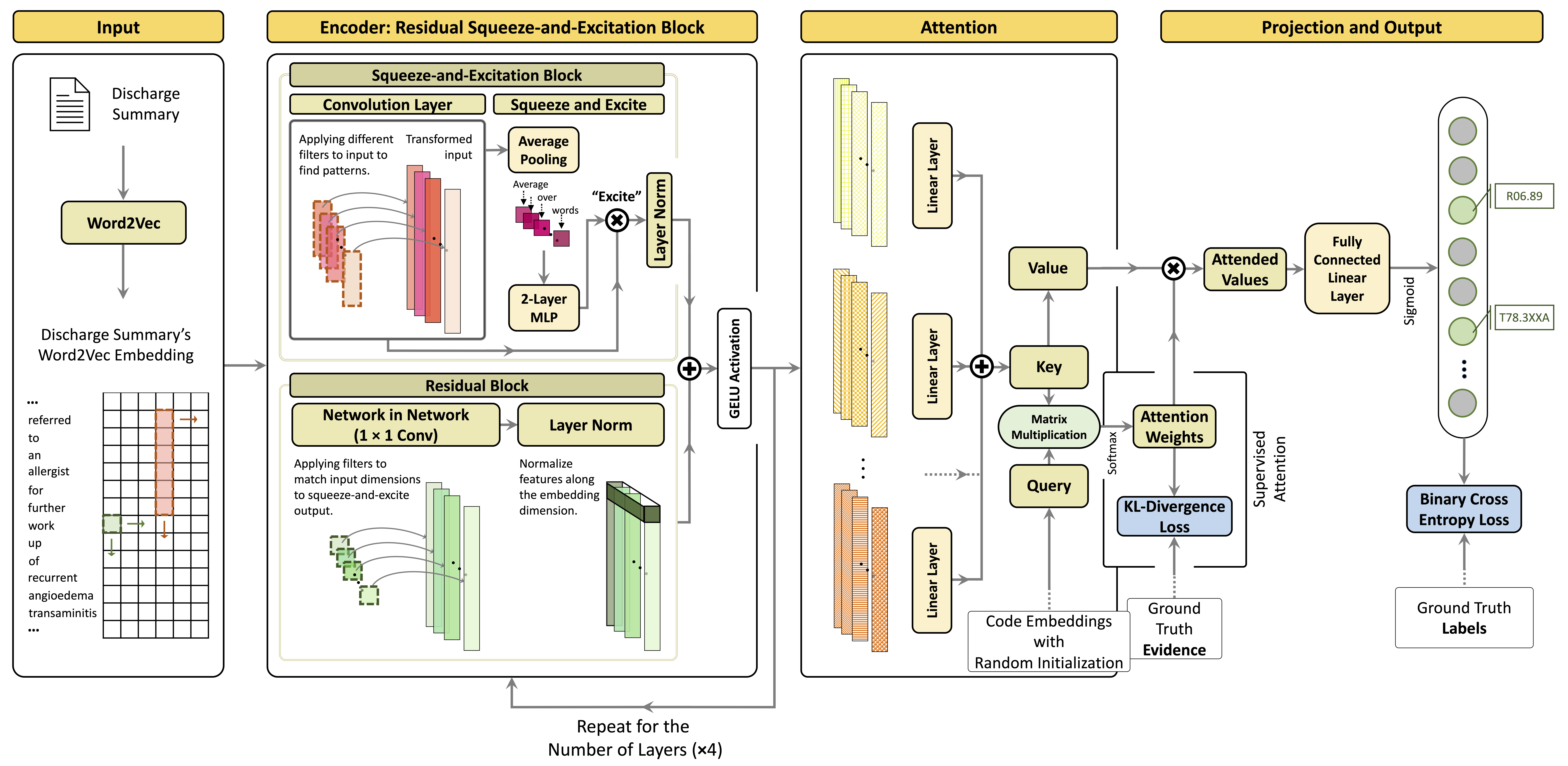}
    \caption{The architecture of EffectiveCAN with supervised attention.}
    \label{fig:architecture}
\end{figure*}

\subsection{Evidence Extraction Methods}

We implemented multiple baseline methods for code evidence extraction, including unsupervised attention, supervised attention, linear tagging, and CNN tagging. Figure \ref{fig:architecture} shows our implementation of the EffectiveCAN model with the attention supervision mechanism for evidence extraction.


\subsubsection{Unsupervised Attention}

EffectiveCAN uses text encoding from multiple layers of Res-SE block to generate the key for the attention module. The result is a label-specific representation of the input obtained by multiplying the key (value) by the attention weights. The attention weights signal the most relevant parts of the input text with respect to the output. Highlighted evidence for predicted codes are tokens whose attention scores are greater than a pre-defined threshold. We consider this the simplest baseline and compare the performance of other supervised methods with it.

\subsubsection{Supervised Attention (SA)}

We added a loss for evidence supervision during training as illustrated in Equation \ref{eq:sup-attn-loss}. We chose Kullback–Leibler (KL) divergence loss over other losses, such as mean squared error, since it is a term in the cross-entropy loss expression and would result in a similar gradient behavior to the binary-cross entropy (BCE) loss used for the code prediction \citep{yu-etal-2021-understanding}.
\begin{align}
    \mathcal{L} = \mathcal{L}_{BCE}(\hat{\textbf{y}}_{code}, \textbf{y}_{code}) + \lambda_1 \, \mathcal{L}_{KLD}(\textbf{a},\textbf{y}_{evd})
    \label{eq:sup-attn-loss}
\end{align}
\noindent where $\textbf{a}$ is the attention weights.

\begin{figure*}[t]
    \begin{center}
    \includegraphics[width=\textwidth]{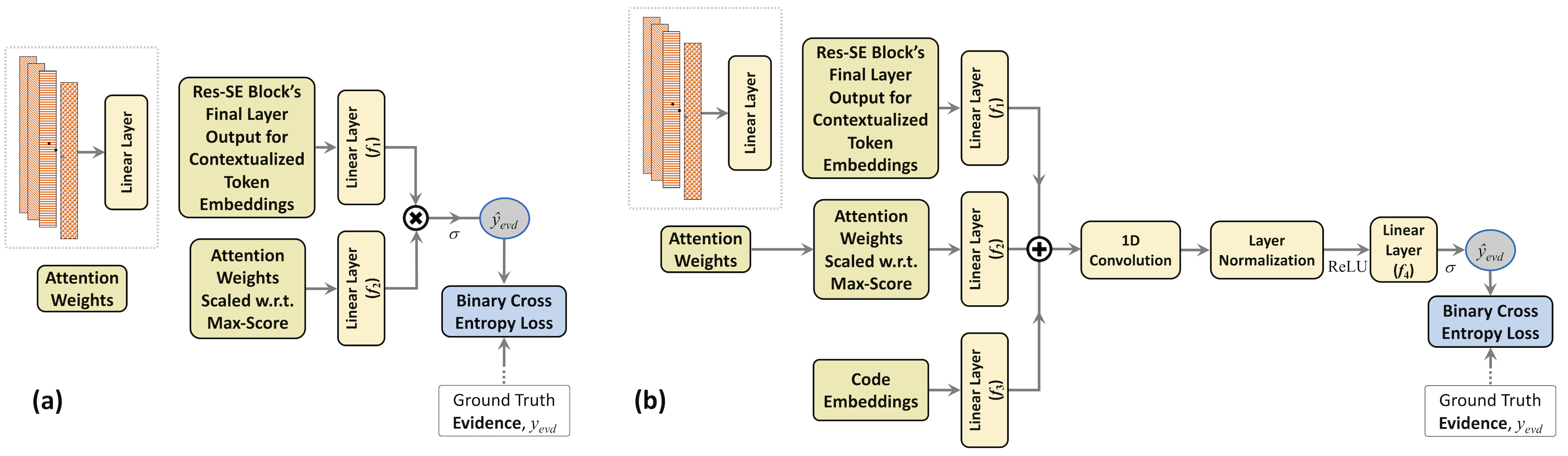}
    \end{center}
    \caption{(a) Linear, and (b) CNN token-level evidence tagging models.}
    \label{fig:evidence-tagging}
\end{figure*}

\subsubsection{Linear Tagging Layer}

Inspired by the work of \citet{segal-etal-2020-simple} on the use of tagging for question answering, we added a feed-forward tagging layer on top of EffectiveCAN for evidence extraction as shown in Fig. \ref{fig:evidence-tagging} (a). We use the output of the last Res-SE block, $\textbf{h}^{l}$, and the normalized attention scores w.r.t. the maximum weight, $\textbf{a}_{scaled}$, as inputs to two linear layers that share parameters for all the labels. The scaling is done so that the maximum score would be consistent among different instances. The outputs of these linear layers are multiplied to obtain the logits for evidence prediction, $\hat{\textbf{y}}_{evd} \in \mathbb{R}^{N}$ (where $N$ is the text length and each token is labeled as evidence or not). We used BCE for the tagging loss, and added it to the label loss through a weight term:
\begin{align}
    \hat{\textbf{y}}_{evd} = \sigma \big(f_{1}(\textbf{h}^{l=4}) \times f_2(\textbf{a}_{scaled}) \big)
    \label{tag-pred}
\end{align}
\begin{align}
    \mathcal{L} = \mathcal{L}_{BCE}(\hat{\textbf{y}}_{code}, \textbf{y}_{code}) + \lambda_2 \, \mathcal{L}_{BCE}(\hat{\textbf{y}}_{evd}, \textbf{y}_{evd})
    \label{eq:tag-loss}
\end{align}

\subsubsection{CNN Tagging Layer}

We extended the linear tagging layer by adding a CNN layer as another method for evidence extraction. The CNN tagger has as input the sum of the two linear projection layers of the last Res-SE block, the normalized attention scores, and the code embeddings, $\textbf{u}$. The inputs are then fed into a 1-D convolutional layer ($\text{conv1D}$) with a kernel size of 9 and out-channel size of 10, followed by layer normalization, ReLU activation, and finally a linear layer ($f_{4}$) to project the output back to the original dimension (see Fig. \ref{fig:evidence-tagging} (b)). 


\begin{align}
    \textbf{x} = f_1(\textbf{h}^{l=4}) + f_2(\textbf{a}_{scaled}) + f_3(\textbf{u})
    \label{cnn-input}
\end{align}
\begin{align}
    \hat{\textbf{y}}_{evd} = \sigma\big(f_4(\text{conv1D}(\textbf{x}))\big)
    \label{cnn-tag}
\end{align}
The output logits from the final layer are used for evidence prediction, with the same BCE loss as the linear tagger, shown in Equation \ref{eq:tag-loss}. 


\section{Experiments and Results}
\label{sec:results}

\begin{table}[t]
\small
    \centering
    \begin{tabular}{lcc}
    \hline
    \textbf{Train Set} & \textbf{Code-F1} & \textbf{Token-F1} \\
    \hline
         $0$            & $58.3$ & $32.0$ \\
         $30 \; (12.5\%)$ & $58.1$ & $32.3$ \\
         $60 \; (25\%)$ & $57.7$ & $32.8$ \\
         $121 \; (50\%)$ & $58.2$ & $33.2$ \\
         $181 \; (75\%)$ & $58.1$ & $36.2$ \\
         $242 \; (100\%)$& $58.1$ & $36.8$ \\
    \hline
    \end{tabular}
    \caption{Supervised attention training performance on dev set for evidence training datasets of different sizes.}
    \label{tab:train-split-size}
\end{table}

\begin{table}[t]
\small
    \centering
    \begin{tabular}{cccc}
    \hline
    \textbf{Data Splits} & \textbf{Train} & \textbf{Dev} & \textbf{Test} \\
    \hline
         Code (c) & c.train & c.dev & c.test \\
          & 47,719  & 1,631 & 3,372 \\
    \hline
         Evidence (ev) & ev.train & ev.dev & ev.test \\
         Inpatient & 181 & 60 & 61 \\
         Profee & 31 & 10 & 11 \\
    \hline
         Code+Evidence & c.train & c.dev & c.test - ev.dev \\
         & + ev.train &  + ev.dev &   - ev.train \\
         Inpatient & 47,900 & 1,691 & 3,131 \\
         Profee & 47,750 & 1,641 & 3,331 \\
    \hline
    \end{tabular}
    \caption{Our new Code+Evidence data splits based on the splits of \citet{mullenbach2018explainable} for code prediction and our evidence dataset splits. } 
    \label{tab:new-splits}
\end{table}

\begin{table*}[t]
\scriptsize
\centering
    \begin{adjustbox}{width=1\textwidth}
    \begin{tabular}{l |c|ccc|ccc} 
    \noalign{\global\arrayrulewidth=1pt}
    \hline
    \noalign{\global\arrayrulewidth=0.5pt}
    \multirow{2}{4em}{Model} & 
    \multirow{2}{4em}{Threshold} & 
    \multicolumn{3}{c|}{Token Match} &  \multicolumn{3}{c}{Position Independent Token Match} \\
    & {} & Precision & Recall & F1 & Precision & Recall & F1 \\
    \noalign{\global\arrayrulewidth=0.5pt}
    \hline
    \textit{CAML}  & & & & & & & \\
    Unsup. Attention &
    $0.05 \pm 0.1$ &
    $17.8 \pm 11.3$   & $27.5 \pm 11.8$  & $21.4 \pm 12.0$ &
    $26.6 \pm 18.6$  & $32.2 \pm 11.1$  & $28.5 \pm 15.4$ \\
    \hline
    \textit{EffectiveCAN} & & & & & & & \\
    Unsup. Attention &
    $0.07 \pm 0.01$ &
    $40.1 \pm 2.3$ & $33.2 \pm 0.6 $ & $36.2 \pm 0.6$ &
    $66.5 \pm 3.8$ & $37.2 \pm 0.4$ & $47.7 \pm 0.8$ \\
    Sup. Attention &
    $0.05 \pm 0.01$ &
    $40.5 \pm 3.0$ & $46.3 \pm 4.1$ & $\mathbf{43.0 \pm 0.2}$ & 
    $65.3 \pm 4.4$ & $50.7 \pm 3.9$ & $\mathbf{56.8 \pm 0.7}$ \\
    Linear Tagging &
    $0.23 \pm 0.06$ &
    $\mathbf{45.6 \pm 1.2}$ & $36.3 \pm 0.8$ & $40.4 \pm 0.1$ &
    $\mathbf{68.8 \pm 1.8}$ & $43.4 \pm 0.4$ & $53.3 \pm 0.8$ \\
    CNN Tagging &
    $0.32 \pm 0.08$ &
    $35.5 \pm 0.4$ & $\mathbf{51.1 \pm 1.4}$ & $41.9 \pm 0.7$ &
    $52.0 \pm 0.3$ & $\mathbf{59.8 \pm 2.0}$ & $55.6 \pm 1.0$ \\
    \noalign{\global\arrayrulewidth=1pt}
    \hline
    \end{tabular}
    \end{adjustbox}
\vspace{0.5pt}
\tiny
\centering
    \begin{adjustbox}{width=1\textwidth}
    \begin{tabular}{l |ccc|ccc} 
    \noalign{\global\arrayrulewidth=1pt}
    \hline
    \noalign{\global\arrayrulewidth=0.5pt}
    \multirow{2}{4em}{Model} & 
    \multicolumn{3}{c|}{Exact Span Match} & 
    \multicolumn{3}{c}{Position Independent Exact Span Match} \\
    & Precision & Recall & F1 & Precision & Recall & F1 \\
    \noalign{\global\arrayrulewidth=0.5pt}
    \hline
    \textit{CAML} & & & & & &  \\
    Unsup. Attention &
    $4.9 \pm 8.2$  & $13.0 \pm 21.2$ & $7.1 \pm 11.8$ &
    $7.7 \pm 12.7$ & $14.5 \pm 23.3$  & $10.1 \pm 16.5$
    \\
    \hline
    \textit{EffectiveCAN} & & & & &  & \\
    Unsup. Attention &
    $19.8 \pm 1.6$ & $35.1 \pm 0.2$  & $25.3 \pm 1.3$ &
    $32.2 \pm 2.3$ & $38.1 \pm 0.1$ & $34.9 \pm 1.4$ \\
    Sup. Attention &
    $20.4 \pm 1.3$ & $\mathbf{44.0 \pm 3.2}$ & $\mathbf{27.8 \pm 0.6}$ & 
    $33.2 \pm 2.5$ & $\mathbf{48.0 \pm 2.6}$ & $\mathbf{39.2 \pm 1.0}$ \\
    Linear Tagging &
    $\mathbf{22.7 \pm 1.0}$ & $34.5 \pm 0.2$ & $27.3 \pm 0.7$ &
    $\mathbf{34.3 \pm 1.6}$ & $41.4 \pm 0.8$ & $37.5 \pm 1.2$ \\
    CNN Tagging &
    $20.0 \pm 0.5$ & $37.9 \pm 1.7$ & $26.2 \pm 0.8$ &
    $29.3 \pm 1.2$ & $46.3 \pm 2.2$ & $35.9 \pm 1.3$ \\
    \noalign{\global\arrayrulewidth=1pt}
    \hline
    \end{tabular}
    \end{adjustbox}
    \caption{Evaluation results of evidence extraction methods on the IP discharge summary test set of \textsc{MDace}.} 
    \label{tab:model-results-ds}
\end{table*}


\begin{table*}[t]
\small
\centering
    \begin{adjustbox}{width=1\textwidth}
    \begin{tabular}{l |c|ccc|ccc|ccc|ccc} 
    \noalign{\global\arrayrulewidth=1pt}
    \hline
    \noalign{\global\arrayrulewidth=0.5pt}
    \multirow{2}{4em}{Dataset} &
    \multirow{2}{4em}{Threshold} &
    \multicolumn{3}{c|}{Token Match} & \multicolumn{3}{c|}{Exact Span Match} & \multicolumn{3}{c|}{P.I. Token Match} & \multicolumn{3}{c}{P.I. Exact Span Match} \\
    & {} & P & R & F1 & P & R & F1 & P & R & F1 & P & R & F1 \\
    \noalign{\global\arrayrulewidth=0.5pt}
    \hline
    Inpatient &
    0.06 &
    37.4 & 37.1 & 37.2 &
    18.0 & 38.1 & 24.5 &
    69.4 & 42.2 & 52.5 &
    34.0 & 42.5 & 37.8
    \\
    Profee &
    0.02 &
    32.6 & 39.4 & 36.5 &
    21.9 & 39.3 & 28.1 &
    41.0 & 41.9 & 41.4 &
    21.1 & 40.4 & 27.7
    \\
    \noalign{\global\arrayrulewidth=1pt}
    \hline
    \end{tabular}
    \end{adjustbox}
    \caption{Evaluation results of the supervised attention model on the code-able notes test set of \textsc{MDace}. } 
    \label{tab:model-results-ip-pf}
\end{table*}

In this section, we describe the experiments for evaluating the evidence extraction methods introduced in Section \ref{sec:methods}, using the token- and span-level metrics in Section \ref{sub-sec:metrics}. 

\subsection{Data Splits}

Rather than simply making random train/dev/test splits, we created sub-training splits to effectively determine the optimum splits for low-resource semi-supervised evidence learning. We randomly sampled fixed development and test sets with $10\%$ of the annotated charts (overall, $20\%$ was held out). Next, we used different portions of the remaining $80\%$ data to create $12.5\%$, $25\%$, $50\%$, $75\%$, and $100\%$ training sets to train the attention weights of the EffectiveCAN model as shown in Table \ref{tab:train-split-size}. As a result, we established the data size needed for supervised training, while the remaining data can be used to create a more representative test set. 

We decided to use the 75\% split point since the evidence training showed only slight improvement with more data. Hence, the created evidence data splits are 60\%/20\%/20\% for train/dev/test. The new data splits for code and evidence are given in Table~\ref{tab:new-splits}\footnote{Four records in the code training set were removed because they do not contain any billing codes.}. We adopted the train/dev splits (c.train and c.dev) of \citet{mullenbach2018explainable} for code prediction as they have been widely used for comparing the performance of deep learning models. We removed the evidence train and dev examples (ev.train and ev.dev) from their test set (c.test) so as to follow the standard data use practices. 

Table \ref{tab:train-split-size} also shows that adding labeled evidence data to the code train/dev sets did not affect code prediction significantly. This is reasonable given that the evidence dataset is much smaller than the code dataset. Compared with the results in \citep{liu-etal-2021-effective}, we can see that the code prediction F1 does not change significantly with or without evidence training. This means that code prediction performance established on the \citet{mullenbach2018explainable} data splits can be transferred to the \textsc{MDace} data splits without much concern.

\subsection{Evaluation Metrics}
\label{sub-sec:metrics}

 We evaluate the evidence extraction methods using the precision, recall, and micro-F1 score on four main metrics: Token match, Exact span match, Position independent (P.I.) token match, and P.I. exact span match. The token match metrics are used to measure the predicted evidence label of each token in a document compared to its ground truth label. The span metrics measure the whole evidence span, which is defined as consecutive tokens with the evidence label. An exact span match considers complete overlap with the ground truth span as correct. These metrics measure how well the evidence extraction methods generate whole spans rather than disjoint, correct tokens. The P.I. metrics disregard the location of the evidence span/token and consider an evidence as correct based on string matching. These metrics are used to alleviate the issue of sufficient vs. complete evidence annotation explained in Section~\ref{sec:ip-pf}. During evaluation, we allow evidence to be generated for all codes regardless of whether or not a code's predicted probability exceeded the prediction threshold. 
 
 We use the model's precision-recall curve on the dev set to determine a threshold that maximizes the token match micro-F1 score, and use this threshold for evaluation on the test set.



\subsection{Results}

The evaluation results of the various evidence extraction methods on the discharge summaries of \textsc{MDace} are shown in Table~\ref{tab:model-results-ds}, obtained by comparing to the ground-truth evidence, irrespective of whether or not the code was predicted. The results for each method/model are from the average of three runs of training. 

Out of all the evidence extraction methods tested, Supervised Attention achieved the best micro-F1 score across all metrics. The tagging methods under-performed SA, likely because they need more data to tune their parameters. The best evidence extraction methods could be based on the size of the training data.

We provide the performance of CAML's attention-based explanation \citep{mullenbach2018explainable} for comparison. It should be noted that 
the best micro-F1 we obtained is $0.523$, lower than the F1 value of $0.539$ as reported in the paper. Additionally, one of the three trained CAML models with different seeds yielded significantly higher evidence performance. As a result, the standard deviation for the reported results is very high.

Since supervised attention resulted in better performance than other methods on discharge summaries, we used it to evaluate the effect of adding other code-able notes including physician and radiology notes to the input (Table~\ref{tab:model-results-ip-pf}). For training the model on Inpatient and Profee datasets, the maximum length for truncating text was increased from $3,500$ to $5,000$. Table~\ref{tab:model-results-ip-pf} shows the performance of Inpatient vs. Profee coding. The position sensitive exact span metrics on Profee are significantly higher than those of Inpatient, likely the result of complete evidence annotations, as the gain disappeared on position-independent metrics. It's worth pointing out that the evidence results on all code-able notes could be affected by input text truncation as potentially more than half of the tokens and evidence were discarded. More experiments and analysis should be conducted to better understand these results.


We determined threshold values based on the token match metric for its simplicity. But we also take into consideration the other metrics, such as exact span match, to have a better grasp of how well the extracted evidence matches human annotations. 
%
%
Note that position independent token match takes tokens out of their context, which may result in evidence that is not reasonable to humans, e.g., ``hr'' where it means hour instead of heart rate. 

We sampled 50 evidence output of the supervised attention model from the Inpatient test set for detailed analysis. We observed that the model was better at extracting short, i.e., single token, evidence (e.g., "hypotension" and "asthma") than evidence with multiple tokens (e.g., "peptic ulcer disease"). Using the Exact span match metric, the SA model predicted 30 (90.9\%) of the 33 short evidence correctly but only 3 (17.6\%) of the 17 multi-token evidence correctly. Although the model couldn't extract the exact multi-token spans, it often identified partial evidence. For example, it generated "peptic" instead of "peptic ulcer disease", and "compartment" instead of "compartment syndrome of left lower extremity". Table~\ref{tab:evidence-examples} in Appendix~\ref{sec:evidence-examples} provides more example outputs from two baseline extraction methods.

Appendix~\ref{app:params} describes the model parameters used for reporting the results.


\section{Conclusions}

In this paper, we introduce \textsc{MDace}, the first publicly available code evidence dataset built on a subset of the MIMIC-III clinical records. The dataset contains evidence spans for diagnosis and procedure codes annotated by professional medical coders. 
\textsc{MDace} addresses a critical need for CAC research to be able to automatically evaluate the code evidence generated by ML models. To the best of our knowledge, \textsc{MDace} is also the only publicly available dataset with evidence annotations for long documents in an extreme multi-label classification setting.

The need for improving the interpretability of text classification models has increased in recent years as they become more complex and opaque. However, datasets with label evidence are rare as the evidence annotations do not occur naturally, nor is the evidence actually used in the real world in those domains, e.g. the rationale annotations on the IMDB reviews \cite{zaidan2007, Pang-Lee-2004}. Recruiting human subjects, especially domain experts, to create an evidence dataset is an expensive and time consuming process. In addition, many applications require the models to be able to generate local explanations \cite{Nguyen-2018}. \textsc{MDace} is a step toward filling the void and can be used to evaluate and enhance the explainability of DL models.
We believe that its release will greatly improve the understanding and application of deep learning technologies for medical coding and text classification.

Given the recent release of the MIMIC-IV clinical notes, our next step is to combine the \textsc{MDace} annotations with the MIMIC-IV dataset and establish baseline performance for ICD-10 code prediction and code evidence extraction.

\newpage
\clearpage

\section{Limitations}

Professional coders are trained to find sufficient, as opposed to exhaustive, evidence for each code. Our Profee coders were instructed to find all the evidence for each code. However, given the large number of notes in some MIMIC encounters, they might only manage to annotate most of the evidence. For Inpatient, there might be more bias among coders towards finding sufficient evidence: namely, there were many cases in which one coder found evidence that another had not, but during the adjudication process, both coders agreed it should be included. Thus, although we have opened the door to automatic evaluation of evidence extraction systems, some metrics, such as recall on our dataset, might underestimate the true recall of a system.

We observed inconsistencies and human errors while cleaning up the data. Coders sometimes only annotated partial evidence, leaving out modifiers like "acute", "moderate" and "bilateral". For example, we consider "bilateral pleural effusions" as the correct evidence but only "effusions" was highlighted, and for "weakness in his lower extremities", only "weakness" was highlighted. Another source of error is due to the limitation of the annotation tool which does not support highlighting and linking discontinuous spans of text as a single evidence for a code. As a result, some evidence may contain extra tokens between the correct evidence tokens and others may miss part of the evidence when the supporting text spans are far apart. We tried our best to fix these issues, but some errors likely remain in the dataset. 

\bibliography{anthology, reference}

\newpage
\clearpage

\appendix

\section{Medical Coding Terminology}
\label{app:code-term}
Medical coding is the process of assigning codes that specify the diagnoses and procedures performed on patients during a visit to a medical facility. For most patient encounters, only a few codes are chosen from the tens of thousands of ICD, CPT, or other procedure codes. Even with pre-defined coding guidelines, there are often significant variations in code selection as medical coding depends on the coder's interpretation. There are two major categories of medical coding: inpatient and outpatient.

Inpatient coding is the coding process applied to documentation created during a patient visit to a medical facility such as a hospital. These admissions are typically for an extended period of time where a variety of tests and procedures are run on the patient. As a result, inpatient records are often long and complex, requiring an experienced medical coder to handle the coding process. Inpatient coding uses two types of code families when assigning codes: ICD diagnosis (CM) and procedure (PCS) codes.

Outpatient coding is the coding process applied to documentation created during shorter patient visits where the patient stay lasts less than 24 hours. The shorter stay typically makes the outpatient coding process simpler and requires fewer codes per encounter than inpatient coding. Outpatient coding includes two types of coding services: professional fee coding (Profee) and facility coding. Profee refers to coding and billing covering the work and reimbursement received by the healthcare provider. Facility coding is the coding and billing for the facility (e.g. hospital or nursing care). Outpatient coding uses CM and current procedural terminology (CPT) codes when assigning codes.

\section{Code Mapping Procedure}
\label{app:codeMap}

Procedure for backward mapping from ICD-10 to ICD-9:

\begin{enumerate}[itemsep=0pt]
    \item Use the identical match or single approximate match from an ICD-10 to ICD-9 code;
    \item When more than one mapping exists, choose the ICD-9 code that is in the MIMIC-III code set. If none of the mapped codes is in MIMIC, choose the code with the description that overlaps the most with that of the ICD-10 code;
    \item When no mapping exists, use the mapped ICD-9 code of the parent ICD-10 code.
\end{enumerate}

This process allows all annotated ICD-10 codes to be mapped except for two in our dataset.

\section{Examples of Initial Disagreement}
\label{app:disagree}

We observed two sources that accounted for the low initial agreement. One source is that the coders annotated the same or similar evidence from different locations of the same documents or in different documents of the same chart. For example, two coders annotated G60.8 for “idiopathic generalized neuropathy”, one from the Physician Initial Consult Note, while the other from the Physician Surgical Admission Note. Both notes are valid for coding. Another example is that one coder assigned I46.9 for “Asystole” documented in the Discharge Summary while the other assigned the same code for “cardiac arrest” from the Physician Initial Consult Note. Both diagnosis terms are correct for I46.9. These cases were resolved in the re-review process, and should be treated as agreements. 


For Profee coding, the initial disagreement was also due to the lack of experience of one coder. An example is that one coder assigned the code S04.40XA for “traumatic 6th nerve palsy” documented in the Discharge Summary whereas the other assigned the code H49.20 for the same diagnosis which is incorrect. The disagreement was resolved after discussion and it was agreed that S04.40XA was the correct code.

\section{Model Parameters}
\label{app:params}

For results given in Table \ref{tab:train-split-size}, $\lambda_1 = 0.5$ was used as the hyperparameter in Equation \ref{eq:sup-attn-loss} without any hyperparameter tuning. The $\lambda$ values in the loss Equations \ref{eq:sup-attn-loss} and \ref{eq:tag-loss} were tuned such that the micro-F1 value for the code prediction task would remain close to the baseline value. For SA, $2.5$ and $5.0$ were considered for the $\lambda$ coefficient, and $\lambda_1=2.5$ yielded code micro-F1 of $0.585$, close to the baseline value of $0.584$. For the tagging models, three values, $0.5$, $1.0$ and $2.0$, were considered, and $\lambda_2=0.5$ yielded code micro-F1 of $0.583$, close to the baseline performance for CNN tagging. These values were used for the reported results. For evidence prediction threshold, steps of 0.02 and 0.05 were used to generate the precision-recall curve for the attention-based and tagging methods respectively, and the threshold values are reported in Tables \ref{tab:model-results-ds} and \ref{tab:model-results-ip-pf}.

The EffectiveCAN based models have about 17 million parameters, and each took about six hours to train on a single NVIDIA Tesla V100 16GB GPU with CO\textsubscript{2} emission of about 680g.

\section{Annotation Guidelines}
\label{app:guide}

The task is to annotate MIMIC charts with sufficient code evidence based on the documentations using an open source tool called INCEpTION.  

\begin{itemize}
    \item For Inpatient coding, annotate evidence for ICD-10-CM and ICD-10-PCS codes.
    \item For Profee coding, annotate evidence for ICD-10-CM and CPT codes (ignoring EM codes which are in the range of 99201-99499).
\end{itemize}

Reference the latest coding book to decide whether an ICD-10 code is supported by the documentation. If there is a definitive diagnosis, do not code symptom codes, otherwise symptom codes can be coded. Code external cause codes only with injury codes.

Code as in real life, once a condition is confirmed and you feel comfortable with a code assignment, annotate the text spans with the code and move on to the next one. You are encouraged to provide multiple evidence for a code, as long as it doesn't slow you down too much. For Profee coding, go through all notes and annotate as many diagnoses as possible.

The general annotation process includes:

\begin{itemize}
    \item Leaf through chart documents to find the ones appropriate to code from.
Highlight best/sufficient text spans as evidence for a code.
    \item Choose the appropriate ICD-10/ICD-9 code pair or CPT code in the Label box to assign to the highlighted text span.
    \item If the correct ICD-10 or CPT code is not in the label set, type it in the Label box and assign it to the highlighted text span.
    \item Try to annotate evidence for all ICD-9 or CPT codes in the label set if there is supporting documentation.
\end{itemize}

Follow these instructions to annotate and export a chart in INCEpTION:

\begin{enumerate}
    \item Go to Dashboard and click Annotation, select a document to open.
    \item Select Search in the left panel. You can search any phrase and select the document containing the phrase to annotate.
    \item Open the Preferences popup, and set the following (Done once for a project):
        \begin{itemize}
            \item Editor: brat (line-oriented)
            \item Sidebar right: 30
            \item Page size: 1000
        \end{itemize}
    \item In a document, double click on a word or highlight a text span, and then select a label from the right panel. You can also start typing in the label box and the matching labels will show up.
    \item You can navigate through the documents using the icons at the top of the middle panel, and move through the annotations using the arrows in the right panel.
    \item After you finish annotating a chart, select Administration -> MIMIC-encounterID -> Settings -> Export, choose WebAnno TSV v3.3 format and then Export the whole project.
\end{enumerate}

These code evidence annotations will be made available to the research communities so those with access to the MIMIC dataset can use them to evaluate the code evidence generated by their ML models.

\newpage
\clearpage

\section{Examples of Generated Evidence}
\label{sec:evidence-examples}
Examples of predicted evidence, using unsupervised attention weights as the baseline and the supervised attention method, are given in Table \ref{tab:evidence-examples}. 

\begin{table}[h]
    \centering
    \begin{adjustbox}{width=1\textwidth}
    \begin{tabular}{l l l l l}
        \textbf{Code} &  \textbf{Human Annotation} & \textbf{Unsupervised Attention} & \textbf{Supervised Attention} & \textbf{Code description}\\
        \hline
        \noalign{\global\arrayrulewidth=1pt}
        \multirow{2}{4em}
        {36.15} & ``left internal mammary artery to & ``mammary'' & ``left internal mammary'' & {Single internal mammary-coronary} \\
        {} & left anterior descending artery'' & {} & ``left anterior descending'' & {artery bypass} \\
        \hline
        \multirow{2}{4em}
        {427.31} & ``Atrial fibrillation'' & ``Atrial'' $\times$ 2 & ``Atrial fibrillation'' $\times$ 2 & {Atrial fibrillation} \\
        {} & {} & ``atrial'' & {} & {} \\
        \hline
        \multirow{2}{4em}
        {424.1} & ``aortic stenosis'' & ``Sj'' & ``Sj'' & {Aortic valve disorders} \\
        {} & {} & {} & ``Aortic (aortic $\times$ 2)'' & {} \\
        \hline
        \multirow{2}{4em}
        {441.2} & ``thoracic aortic aneurysm'' & ``thoracic'' & ``thoracic'' & {Thoracic aneurysm without mention} \\
        {} & {} & {} & ``aneurysm'' & { of rupture} \\
        \hline
        \multirow{1}{4em}
        {428.0} & ``Congestive heart failure'' & ``Congestive'' $\times$ 2 & ``Congestive heart failure'' $\times$ 2 & {Congestive heart failure, unspecified} \\
        \hline
        \multirow{2}{4em}
        {790.92} & ``Supratherapeutic INR'' & ``INR'' $\times$ 3 & ``INR'' & {Abnormal coagulation profile} \\
        {} & {} & {} & ``Supratherapeutic INR'' & {} \\
        \hline
        \multirow{3}{4em}
        {584.9} & ``Acute Renal Failure'' & ``Renal''  & ``Acute Renal Failure'' & {Acute kidney failure, unspecified} \\
        {} & {} & ``creatinine'' & ``renal failure'' & {} \\
        {} & {} & ``renal'' & {} & {} \\
        \hline
        \multirow{1}{4em}
        {585.9} & ``Chronic renal insufficiency'' & ``renal'' $\times$ 2  & ``renal insufficiency'' $\times$ 2 & {Chronic kidney disease, unspecified} \\
        \hline
    \end{tabular}
    \end{adjustbox}
    \caption{Examples of generated evidence}
    \label{tab:evidence-examples}
\end{table}

\newpage
\clearpage

\end{document}